%% file: acl2020.tex
\documentclass[11pt,a4paper]{article}
\usepackage[hyperref]{eacl2021}
\usepackage{helvet}
\usepackage{courier}
\usepackage{times}
\usepackage{latexsym}
\usepackage{url}
\usepackage{graphicx}
\usepackage{tabularx}
\usepackage{booktabs}
\usepackage{amsmath}
\usepackage{amsfonts}
\usepackage{afterpage}
\usepackage{subfig}
\usepackage{float}
\usepackage{verbatim}
\usepackage{xcolor}
\usepackage{amssymb}
\usepackage{pifont}
\usepackage{colortbl}

\usepackage{microtype}
\usepackage{multirow}

\aclfinalcopy 
\def\aclpaperid{} 

\usepackage{fdsymbol}
\newcommand\ai{$^\diamondsuit$}
\newcommand\uw{$^\spadesuit$}
\newcommand\msr{$^\clubsuit$}
\newcommand\stan{$^\vardiamondsuit$}
\newcommand\cape{$^\heartsuit$}

\usepackage{humanist}
\usepackage[T1]{fontenc}
\newcommand{\model}{{Co-opNet}}
\newcommand{\dataset}{{ArXiv}}

\definecolor{blue}{RGB}{98, 75, 153}
\definecolor{Gray}{gray}{0.95}
\newcolumntype{a}{>{\columncolor{Gray}}r}
\newcommand{\cmark}{\ding{51}}%
\newcommand{\xmark}{\ding{55}}%

\title{Discourse Understanding and Factual Consistency \\ in Abstractive Summarization}

\author{
Saadia Gabriel\uw \hspace{10pt}
Antoine Bosselut\stan \hspace{10pt} 
Jeff Da \ai \hspace{10pt} 
Ari Holtzman\uw \hspace{10pt} \\
\textbf{Jan Buys}\cape \hspace{10pt} 
\textbf{Kyle Lo} \ai \hspace{10pt} 
\textbf{Asli Celikyilmaz} \msr \hspace{10pt} 
\textbf{Yejin Choi}\uw\ai\\ 
  \uw Paul G. Allen School of Computer Science \& Engineering,
  University of Washington \\ 
  \msr Microsoft Research 
  \ai Allen Institute for Artificial Intelligence \\
  \stan Stanford University 
  \cape University of Cape Town \\ 
  \tt \normalsize {\{skgabrie,ahai,yejin\}@cs.washington.edu, \{jeffd,kylel\}@allenai.org}, \\ \tt \normalsize antoineb@cs.stanford.edu, jbuys@cs.uct.ac.za, aslicel@microsoft.com 
}

\date{}

\begin{document}
\maketitle
\begin{abstract}

We introduce a general framework for abstractive summarization with factual consistency and distinct modeling of the narrative flow in an output summary. Our work addresses current limitations of models for abstractive summarization that often hallucinate information or generate summaries with coherence issues. 


 To generate abstractive summaries with \emph{factual consistency} and \emph{narrative flow}, we propose \emph{Cooperative Generator -- Discriminator Networks} (\model), a novel transformer-based framework where a generator works with a discriminator architecture to compose coherent long-form summaries. We explore four different discriminator objectives which each capture a different aspect of coherence, including whether salient spans of generated abstracts are hallucinated or appear in the input context, and the likelihood of sentence adjacency in generated abstracts. 
 
We measure the ability of \model~to learn these objectives with arXiv scientific papers, using the abstracts as a proxy for gold long-form scientific article summaries. 
Empirical results from automatic and human evaluations demonstrate that \model~learns to summarize with considerably improved global coherence compared to competitive baselines. 

\end{abstract}

\input{1-intro.tex}
\input{2-model.tex}
\input{3-dataset}
\input{5-experiments}
\input{7-related.tex}

\input{8-conclusion}
\input{9-acknowledgments}
\bibliographystyle{acl_natbib}
\bibliography{tacl2018}
\clearpage
\input{appendix}

\end{document}

%% file: 1-intro.tex
\section{Introduction}

\begin{figure}
    \centering
    \includegraphics[width=.8\linewidth]{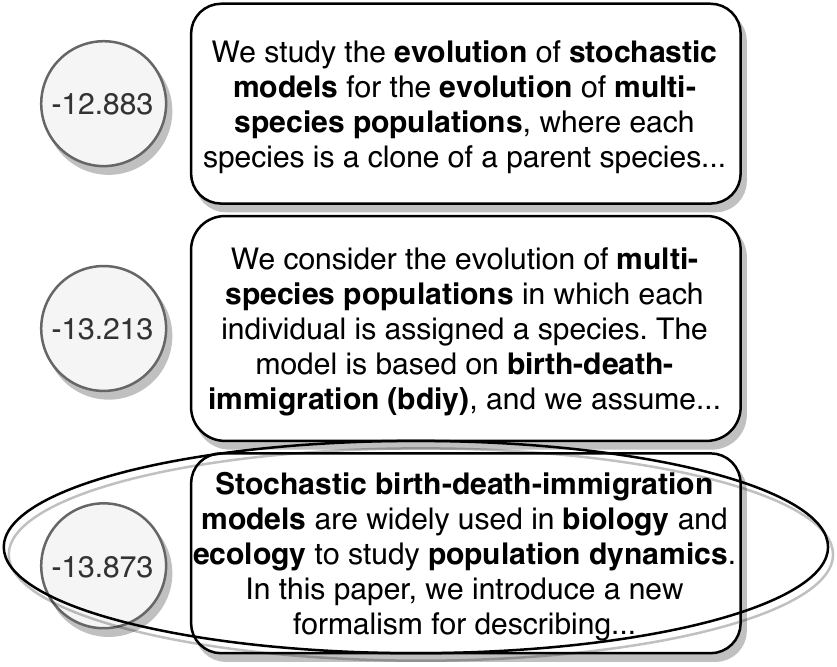}
    \caption{Generated abstracts for a biology article (from the \textbf{Bio} subset of our arXiv dataset). Abstracts are ranked from most (top) to least likely (bottom) using the generator model. 
    Abstracts with better narrative structure and domain-specific content (such as the circled abstract) are often out-ranked in terms of likelihood by abstracts with factual errors and less structure.}
    \label{fig:cands}
\end{figure}


Generating summaries with coherent discourse structure and domain knowledge awareness poses a challenge for current methods in summarization. Generative models can commonly produce high quality text (Figure \ref{fig:cands}), but fail to understand finer-grained details of coherence such as the structure and flow of a narrative. In addition, they often generate factually incorrect content. Prior work on factuality in abstractive summarization has found that current models can hallucinate information more than 70\% of the time when generating summaries of news articles \cite{maynez2020faithfulness}. 

To address these issues, we focus our study on generating abstractive summaries with \emph{factuality} and \emph{narrative flow}. Given an input document, the goal is to generate a paragraph-length abstractive summary with proper discourse structure that contains factually correct claims.
Our study builds on and extends previous work that focuses on either  
\emph{extractive document-level} summarization \cite{Nenkova2012ASO,allahyari2017text} or \emph{abstractive sentence-level} summarization \cite{rush2015neural,newsroom,xsum}. 




In pursuit of this goal, we introduce \emph{Cooperative Generator-Discriminator Networks} (\model), a framework for abstractive summarization that considers subtle aspects of fact-checking and discourse necessary for coherent text generation. In this framework, the generator, a transformer language model fine-tuned for abstractive summarization, proposes a pool of candidate summaries ($\S$2).
The discriminator, also transformer-based, scores the factuality or discourse quality of candidate summaries using one of four different objectives: the overlap between a scientific article introduction and predicted fact-checking evidence spans in generated summaries, the ordering of predicted discourse roles, the coverage of predicted discourse roles, or the likelihood of adjacency between generated sentences ($\S$3).
The best summary is chosen cooperatively by combining the generator and discriminator scores ($\S$4). 

Most previous works on \emph{abstractive document-level} summarization have difficulty in directly modeling or evaluating \emph{narrative flow} and \emph{factuality} in generated summaries. This weakness is largely due to the inherent limitations of existing datasets, such as the CNN/DailyMail dataset \cite{cnndm}. The reference summaries available in these commonly used resources are mainly headlines of news articles or stories. As a result, they are often sets of disconnected sentences that are highly extractive, leading to models that are also extractive \citep{Hoang2019EfficientAO}, rather than abstractive.  

%

In order to address these data challenges, 
we test our summarization model on a set of arXiv scientific papers. Scientific abstracts are ideal for modeling narrative flow as they are structured with highly coherent discourse flow. They also maintain implicit \emph{abstractive} alignments with respect to the introduction of the article -- in contrast to the tight,  \emph{extractive} alignments of current models. Scientific article summarization is also a task where factuality is more well-defined than in other domains like story summarization which leave more room for interpretation.


Comprehensive empirical results considering both automatic and human evaluations demonstrate that \model~learns to summarize scientific articles from three domains with considerably improved global coherence compared to competitive baselines ($\S$6). We also demonstrate that the framework is generalizable to multiple coherence objectives, and effective at generating scientific abstracts that are more factually consistent. 

%% file: 2-model.tex
\begin{figure*}[t]
    \centering
    \includegraphics[width=.7\textwidth]{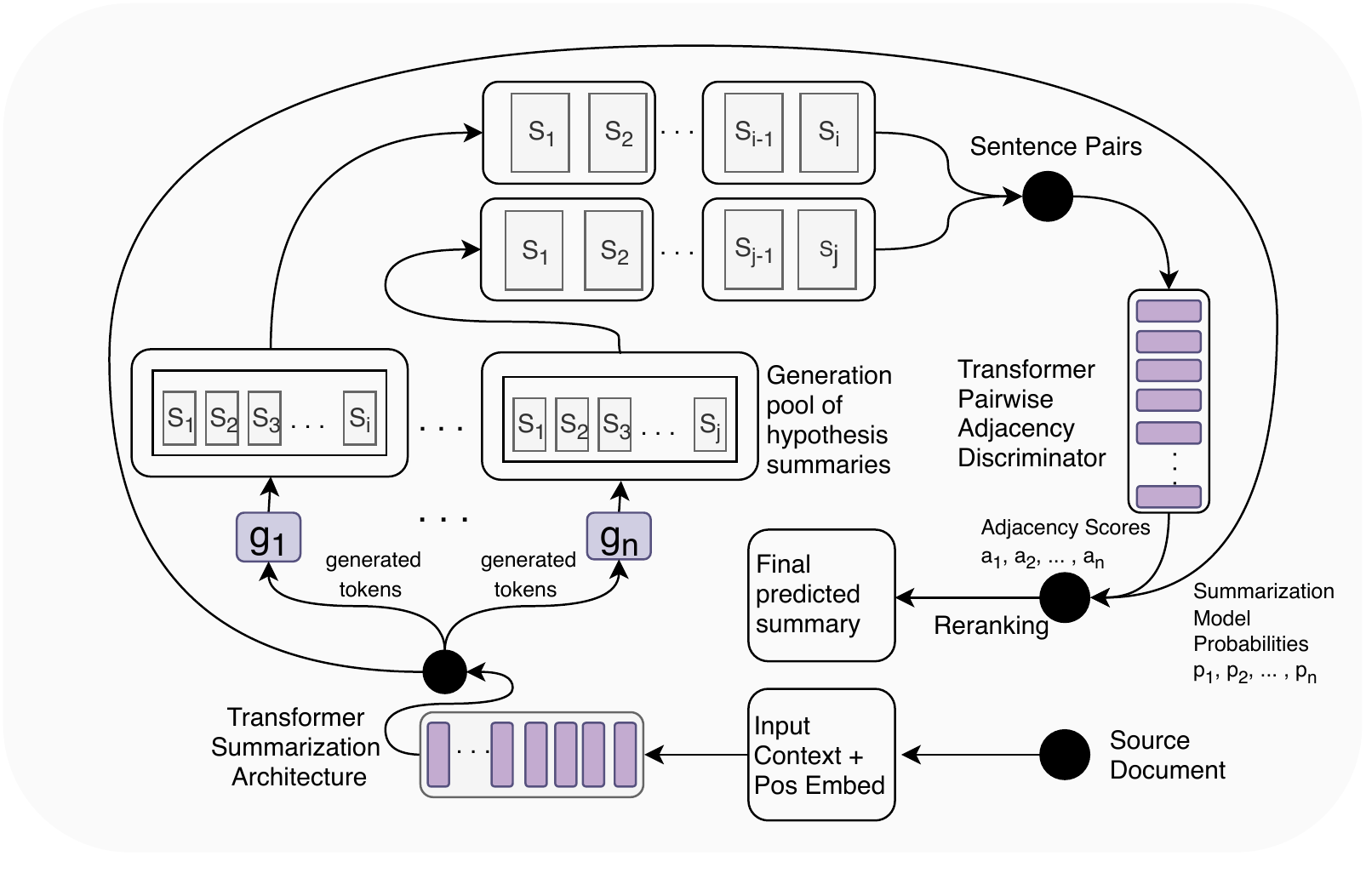}
    \caption{Model architecture for adjacency reranking variation of \model}
    \label{fig:generator}
\end{figure*}

\input{models/generators.tex}
\input{models/disc.tex}

\input{models/cooperation.tex}

%% file: models/generators.tex
\section{Generator Networks}
\label{sec:gen}
We use the transformer architecture of \citet{Radford2019LanguageMA} as our generator's architecture. Following the work of \citet{Liu2018GeneratingWB}, we adapt a language model to the task of abstractive summarization by concatenating the article $a$, a delimiter token $[\mathrm{SEP}]$, the summary $s$, and an end token $[\mathrm{END}]$ into one input vector $X = (a_1,...,a_{\vert a \vert},[\mathrm{SEP}],s_1,...,s_{\vert s \vert},[\mathrm{END}])$, where $\vert a \vert$ is the length of the gold article and $\vert s \vert$ is the length of the gold summary. 

At each time step $i$, the model produces an output probability distribution over the vocabulary for the next token $w_i$ given all previous output tokens $w_{<i}$. For any arbitrary token $w_j$ preceding $w_i$, the per-layer representation of that token is computed in the following way: 
\begin{align}
\textbf{h}_{j}^0 &= \textbf{W}_e(w_{j}) + \textbf{p}_{j} \label{eq:input}\\
\textbf{h}_j^l &= block(\{\textbf{h}\}_{< j}^{l-1})
\label{eq:block} 
\end{align}

where \textit{block} refers to each transformer block composed of multi-headed attention, a feedforward network and layer normalization, $\textbf{W}_e$ is a word embedding matrix, $\textbf{p}_{j}$ is the position embedding, $\textbf{h}_{j}^0$ is the initial representation, $\{\textbf{h}\}_{j}^{l}$ is the block output for an arbitrary layer $l$, and $\{\textbf{h}\}_{< j}^{l-1}$ is the set of all block outputs from the preceding layer for positions up to $j$. Finally, for the current position $i$ in the sequence, we compute a distribution over the output vocabulary as follows:
\begin{align}
P(w_i | w_0,...w_{i-1}) &= \mathrm{softmax}(\textbf{h}_{i-1}^L \textbf{W}_e) \label{eq:prob}
\end{align}
where $\textbf{W}_e$ is the same embedding matrix as in Equation~\ref{eq:input} and $\textbf{h}_{i-1}^L$ is the final layer transformer block output. 

%% file: models/disc.tex
\section{Discriminator Networks} 
\label{sec:disc}
Because summarization models are prone to narrative flow and factual consistency issues \cite{kryciski2019evaluating,Xu2020DiscourseAwareNE}, we use a discriminator to score generated summaries for discourse and factuality properties. 
Due to the challenge of explicitly defining discourse and factuality properties as scores, these properties are approximated using parameterized scoring functions. 

These scoring functions determine if generated text demonstrates discourse and factuality properties in three ways: (1) predicting the discourse role of sentences within a full summary, (2) predicting the likelihood of adjacency given a sentence pair, and (3) measuring the presence of salient facts in the generated summary from the original input context. While our discriminators focus on these three properties, we note that this framework is generalizable and could be extended to include other discriminator models that encourage different communicative norms associated with high-quality language generation. 

\subsection{Discourse}
We explore different discriminator architectures as additional discourse scoring functions during the generator's decoding process.
For these discriminators, we generally score discourse in two ways. First, we use inferred sentence-level scientific abstract discourse role labels\footnote{The labels are \{BACKGROUND, METHOD, OBJECTIVE, RESULT, OTHER\}.} defined by \citet{Cohan2019PretrainedLM}  and predict them using a sequence classifier\footnote{See \cite{Cohan2019PretrainedLM} for model and training details. } based on SciBERT \cite{Beltagy2019SciBERTPC}. Using these predictions, we score the discourse properties of the abstract relative to their coverage (\S\ref{ssec:coverage}) or ordering (\S\ref{ssec:order}). Second, we learn a function that can score the likelihood that sentences within generated abstracts should be adjacent to one another (\S\ref{ssec:adj}). 

\subsubsection{Coverage} 
\label{ssec:coverage}
We measure the completeness of the narrative structure within a scientific abstract by defining the following coverage score: 

\begin{align}
    \mathcal{L}_{cov} &= 
    \text{log}(D_{abs}/D_{all}),
\end{align}

\noindent where $D_{abs}$ is the number of unique discourse roles appearing in an abstract and $D_{all}$ is the total number of possible discourse roles. This objective allows us to penalize abstracts that are missing discourse roles. For example, an abstract that fails to mention anything about the results of the study would be penalized.

\subsubsection{Ordering}
\label{ssec:order}
We also score the order in which discourse labels appear in generated abstracts. In Table \ref{table:dro}, we hard-code valid orderings of discourse labels for generated sentences based on each of the abstract discourse roles of \citet{Cohan2019PretrainedLM}. If the ordering for two adjacent sentences in the abstract $O(s_{i-1}, s_{i})$ is valid, the score for the ordering is 1 (-1 otherwise). We sum the scores for all the orderings within a particular abstract and normalize between 0 and 1 (as described by \text{$f_{n}$}):\footnote{See the Appendix for a more detailed description of the \text{$f_{n}$} function.} 

\begin{align}
    \mathcal{L}_{order} &= 
    \text{log}(\text{$f_{n}$}(\sum_{i=1}^S O(s_{i-1}, s_{i})))
\end{align}

\begin{table}[t]
\centering
\scalebox{0.7}{%
\begin{tabular}{c|c} \toprule
     \multicolumn{1}{c}{$S_{i-1}$} & $S_{i}$  \\ \midrule
     BACKGROUND & BACKGROUND  \\
     BACKGROUND $\lor$ METHOD $\lor$ OBJECTIVE & METHOD  \\ BACKGROUND $\lor$ OBJECTIVE $\lor$ METHOD & OBJECTIVE \\
     OBJECTIVE  $\lor$ METHOD $\lor$ OTHER & RESULT \\   \bottomrule
\end{tabular}}
\caption{Discourse Role Ordering}
\label{table:dro}
\end{table}

We also impose a rule for $s_1$=`BACKGROUND' and a rule for $s_S$=`RESULT' to encourage more natural orderings. 

\subsubsection{Adjacency Classification}
\label{ssec:adj}
To model the likelihood of adjacency between two sentences $s_u$ and $s_v$, 
we first compute a hidden representation of the sentence pair using SciBERT \cite{Beltagy2019SciBERTPC}. 
The encoder input is the concatenation of the sentences: $\textbf{s} = [\mathrm{CLS}] + s_u +  [\mathrm{SEP}] + s_v + [\mathrm{SEP}]$, where $[\mathrm{CLS}]$ is a special token associated with the task and $[\mathrm{SEP}]$ is a sentence delimiter token. 
Each word in the sequence is encoded by a word embedding $w_i$ and positional embedding $p_i$ and passed through the SciBERT model to yield $\textbf{h}_{cls}$, the output state at the position of the $[\mathrm{CLS}]$ token. 
We then obtain the probability of adjacency between the sentences 
by a linear projection of $\textbf{h}_{cls}$ followed by a sigmoid activation: 
\begin{equation}
    P_{adj}(\textbf{s}) = \sigma(\textbf{w}_{disc}^{\top} \textbf{h}_{cls}) \label{eq:p_disc}
\end{equation}


\noindent We define the training objective for the adjacency discriminator to minimize the negative log likelihood of predicting whether two sentences are adjacent or not:
\begin{align}
    \mathcal{L}_{disc} &= 
    -\Big(\delta_{adj}(\textbf{s}) \cdot \log(P_{adj}(\textbf{s})) \notag \\ 
    & + (1 - \delta_{adj}(\textbf{s})) \cdot \log(1 - P_{adj}(\textbf{s}))\Big),
\end{align}


\noindent where $\delta_{adj}(\textbf{s})$ is an indicator function for whether the two sentences in $\textbf{s}$ are adjacent. 
We note that while the discourse discriminators mainly focus on narrative structure, they may also capture context-aware aspects of factuality and content selection.

\subsection{Factuality and Faithfulness}

To measure factuality of generated summaries, we predict which tokens in the summary are likely to belong to a fact-checking evidence span (i.e., a span of the text used to prove a scientific claim 
using a finetuned BERT token classification model.\footnote{See Appendix \ref{disc-b} for details of token classification model.} 
Recent work has shown that inspecting attention weights alone is not necessarily a reliable metric for determining saliency of particular aspects in the input context to the output of neural models \cite{serrano2019attention}. 
The saliency weights representing the likelihood of tokens belonging to evidence spans provides us with a more explicit representation of factual importance. 

We obtain proxy saliency labels for the importance of a particular token $t$ appearing in an abstract 
using a BERT model trained on evidence spans annotated for scientific fact-checking \cite{wadden2020fact}. Specifically, if $t$ is not a stopword and $t \in E$, where $E$ is an evidence span used to check a scientific claim, then we assign a label of 1 to $t$. Otherwise, the label for $t$ is 0. Examples of extracted spans are given in table \ref{table:fact}. 

We compare the predicted evidence spans against information presented in the original introduction to capture the degree to which generative models are hallucinating information.  

 \begin{table}[t]
 \centering
 \resizebox{\columnwidth}{!}{%
 \begin{tabular}{  l  c } 
 \toprule
 \textbf{Topic}  & \textbf{Spans} \\
 \midrule
   \multirow{2}{*}{\textbf{NLP}} &  existing semantic schema, annotation effort,  \\ 
   & music knowledge representation, siri assistant  \\ \midrule
   \multirow{2}{*}{\textbf{BIO}} & biological system, ptotic, cybernetics \\
   & entropy , shannon established fundamental limits \\
 \bottomrule
 \end{tabular}
 }
 \caption{Salient spans extracted using factuality discriminator.}
 \label{table:fact}
 \end{table}

\paragraph{Factuality Objective} 

At inference time, we compare the extracted salient spans, $F(g)$, of the generated summary $g$ against the set of all ngrams in the article input context, $N(a)$, measuring the degree to which salient spans are hallucinated:
\begin{align}
    \mathcal{L}_{fact} &= 
    \text{log}(\frac{|\{f | f \in F(g), f \in N(a) \}|}{|F(g)|})
\end{align}


%% file: models/cooperation.tex
\section{Reranking with Discourse and Factuality Experts}
\label{sec:cooperation}

To incorporate the discriminator objective into our summarization framework, 
we first generate a pool of candidate summaries from the base summarization model (\S2) 
using any decoding strategy (e.g., beam search or top-$k$ sampling). Then, the discriminator is used to re-rank these candidates in conjunction with the original token-level generator scores.
For example, in the case of the adjacency discriminator, we maximize the generator token-level probability of a candidate summary $g$, 
and the average of adjacency scores for the set of sentences composing $g$ (denoted $S(g)$) -- i.e., the probability of each sentence $s_u$ being adjacent to the previous sentence $s_{u-1}$ in $S(g)$:
\begin{align}
    \mathrm{score}(g) &= \lambda_{gen}\frac{1}{|g|}\sum_{i=1}^{\vert g \vert} \log P(w_i|w_1,...w_{i-1}) \notag \\ 
&+ \lambda_{disc}\frac{1}{|S(g)|-1}\sum_{u=2}^{|S(g)|} \log P_{adj}(s_u, s_{u-1}), \label{eq:final_gen_score}
\end{align}
\noindent where $\lambda_{gen}$ and $\lambda_{disc}$ are hyper-parameters controlling the contribution of the generator and adjacency discriminator to the final predicted summary. The same procedure is followed for the other discourse and factuality objectives, replacing $P_{adj}(s_u, s_{u-1})$ with the scores from these discriminators.  



%% file: 3-dataset.tex
\section{Data}


\subsection{Datasets}



Since the focus of this work is on generating summaries with more coherent narrative flow and greater factual consistency, we concentrate on datasets requiring discourse structure to generate good summaries. Particular attributes of the discourse structure of these datasets include: 
\begin{itemize}
    \item Length of summaries $\rightarrow$ Are the summaries long enough to clearly show narrative flow properties and factual correctness?
    \item Abstractiveness of gold summaries $\rightarrow$ Do the summaries exhibit particular sentence-level flow, or are the summary sentences extracted highlights from the context?
\end{itemize} 


\paragraph{ArXiv} We crawled over 700K samples (472K abstracts) from scientific articles on \texttt{arxiv.org}. In our experiments we primarily focus on the CS\footnote{\url{https://arxiv.org/corr}} and Bio\footnote{\url{https://arxiv.org/archive/q-bio}} domain subsets. The task we define is to generate an abstract given a introduction, which presents a challenge to existing summarization models. This task also requires models to learn relevant domain knowledge for the scientific domain of interest and recognize common discourse structure for papers written in that domain.  


 \begin{table}[t]
 \centering
 \resizebox{.7\columnwidth}{!}{%
 \begin{tabular}{  l  r r r r r } 
 \toprule
 \textbf{Split}  & \textbf{CS} & \textbf{BIO} & \textbf{AAN}  \\
 \midrule
 Train & 44900 & 4104 & 10106      \\ 
 Validation  & 5622 & 555 & 892  \\
 Test  & 5670 & 522 & 892 \\
 \bottomrule
 \end{tabular}
 }
 \caption{Domain subset sizes}
 \label{table:domain}
 \end{table}

\paragraph{AAN}
Additionally, we include an existing dataset of scientific articles that focuses on papers in the NLP computer science domain. This dataset consists of a 12k paper subset from the ACL Anthology Network (AAN; \citealp{Radev2009TheAA}) with extracted introduction and abstract pairs.
\\
\\
Scientific abstracts in \dataset~and AAN have properties that are missing from existing summarization datasets based on Newswire data. For example, XSum \cite{xsum} and Newsroom \cite{newsroom} summaries are generally too short to exhibit cross-sentence narrative flow. Meanwhile, CNN/DailyMail \cite{cnndm} summaries are acquired by concatenating extracted highlights, which can be unrelated. Conversely, \dataset~and AAN abstracts are long enough to have multiple sentences,\footnote{See Appendix \ref{datasets} for comparison of datasets.} and generally exhibit strong discourse patterns typical to scientific writing, making them ideal corpora for assessing discourse understanding in abstractive summarization. Table \ref{table:domain} provides details of dataset splits.




%% file: 5-experiments.tex
\section{Experimental Setup}


\begin{table*}[!ht]
\centering
\scalebox{0.8}{%
\begin{tabular}{l|rrr|rrr|rrr} \toprule
\multicolumn{1}{l}{\multirow{2}{*}{\textbf{Model}}} & \multicolumn{3}{c}{\textbf{AAN}} & \multicolumn{3}{c}{\textbf{CS}} & \multicolumn{3}{c}{\textbf{Bio}}\\
\multicolumn{1}{c}{}& \textbf{R-1} & \textbf{R-2} & \multicolumn{1}{c}{\textbf{R-L}} & \textbf{R-1} & \textbf{R-2} & \multicolumn{1}{c}{\textbf{R-L}} & \textbf{R-1} & \textbf{R-2} & \multicolumn{1}{c}{\textbf{R-L}} \\
\midrule
Lede-3  & 27.12 & 6.62  & 23.88 & 28.22 & 7.06  & 16.22 & 27.60 & 5.70  & 24.21 \\
LexRank & 36.03 & 10.14 & 31.37 & 36.53 & 10.41 & 32.09 & 35.32 & \textbf{8.84}& 30.76 \\
LSTM    & 27.80 & 5.57 & 18.02 & 22.74 & 4.56 & 20.64 & 10.73 & 0.49 & 9.94 \\
PGen & 39.85 & 12.83 & 23.24 & 36.68 & \textbf{11.74} & 32.55 & 23.74 & 4.48 & 21.65 \\ \midrule
Generator (Our work) & 41.31  & \textbf{12.97}  & 37.05  & 38.01  & 10.95  & 34.46  & 34.86  & 8.45   & 31.38  \\ 
\model~(Our work) & \textbf{41.67} & 12.65 & \textbf{37.23} & \textbf{38.57} & 10.81 & \textbf{35.11} & \textbf{35.86} & 8.41 & \textbf{32.56} \\
\bottomrule
\end{tabular}}
\caption{Automatic Evaluation of generative architectures and Co-opNet. For AAN, we provide results using the Factuality discriminator. For CS and Bio, we provide results using the Coverage discriminator.}
\label{table:auto_gen}
\end{table*}


\begin{table}[!ht]
\centering
\scalebox{0.8}{%
\begin{tabular}{lrr} \toprule
     \multicolumn{1}{l}{\textbf{Model}} & \multicolumn{1}{c}{\textbf{BERTScore}} & \textbf{SciBERTScore}  \\ \midrule
     {PGen}      & 57.86 & 59.13  \\
     {Generator} & 61.71 & 62.80  \\ \midrule
     {Co-opNet (Adj)}  & 61.87 & 63.10 \\
     {Co-opNet (Fact)} &  \textbf{62.09} & \textbf{63.21} \\
\bottomrule
\end{tabular}}
\caption{BERTScore results on AAN subset (F1) }
\label{table:auto_bert}
\end{table}

Our implementation is based on the Huggingface implementation\footnote{\url{ https://github.com/huggingface/transformers}} of the BERT \citep{Devlin2018BERTPO} and GPT-2 language models \citep{Radford2019LanguageMA}. 

\paragraph{Generator} 

We perform WordPiece tokenization for the input context and output summaries. Because of the fixed input size of the transformer language model, the input context is truncated to a maximum of 800 tokens, and summaries are truncated to a maximum of 200 tokens. We use a learning rate of 2e-5 and a batch size of 16 to finetune the generator. We train the base summarization transformer model for 12 epochs. All experiments are run on either a Titan-X or Quadro RTX 8000 GPU. Training time for the AAN and \dataset~Bio datasets is about 30 minutes per epoch. Training time for the \dataset~CS dataset is 2.5 hours per epoch. In our experiments we use top-$k$ sampling with $k$=4 \cite{Fan2018HierarchicalNS} 
to generate candidate summaries for each model.

\paragraph{Discriminator} 

At training time we use a maximum sentence length of 200 tokens to accommodate the fixed input size of BERT (512 tokens), reduce inference time, and discourage the model from generating abnormally long run-on sentences that indicate the presence of coherence issues.\footnote{See the original papers for details of training the SciFact and abstract discourse models.} 

For the adjacency discourse models, we fine-tune the discriminator using a learning rate of 2e-5, a linear warmup learning rate schedule, and a batch size of 32. All adjacency discourse discriminator models are fine-tuned for 2 epochs on a Titan-X GPU. The adjacency discriminator models are adapted from the Huggingface implementation of the BERT next sentence prediction classifier. We initialize the 12-layer BERT-base discriminator model with the pretrained weights of the SciBERT-uncased model, which was originally trained on 1.14 million scientific papers \cite{Beltagy2019SciBERTPC}. Two discriminators are trained: one is fine-tuned on AAN for decoding both \dataset~CS and AAN, while the other discriminator is fine-tuned on \dataset~Bio and used exclusively for decoding that subset.
We weigh the generation and discriminator models equally when decoding by setting $\lambda_{gen}$=$\lambda_{disc}$=$.5$. Additional implementation details are provided in Appendices \ref{disc-a} and \ref{disc-b}.\footnote{Our code/data is released here: \url{https://github.com/skgabriel/coopnet}.} 


\section{Experiments}

We compare against extractive approaches using the Lede-3 and LexRank \cite{Erkan2004LexRankGC} baselines. We also compare against two abstractive approaches: a 2-layer bi-LSTM sequence-to-sequence model with attention (LSTM), and a pointer-generator model (PGen; \citealp{see2017get}). Training details of the supervised baselines can be found in the Appendix \ref{baselines}. In addition, we compare to a subset of our approach that only uses the generator to produce summaries, rather than the full framework.

\input{experiments/automatic_eval.tex}

\input{experiments/human_eval.tex}

%% file: experiments/automatic_eval.tex

\subsection{Automatic Evaluation} 
\label{ssec:traditionalautoeval}

 Following previous work on summarization, we use the ROUGE metric \cite{lin2004rouge} for automatic evaluation of generative models and \model. 
Specifically, we report ROUGE-1, ROUGE-2 and ROUGE-L F1 scores. To capture similarity in contextual meaning, we look at BERTScore F1 \cite{Zhang2019BERTScoreET}, which has been shown to more closely correlate with human judgements than other generation metrics.  

Results on the AAN, CS and Bio subsets of \dataset~are shown in Table \ref{table:auto_gen}. \model{} outperforms all baselines on ROUGE-1 and ROUGE-L by a consistent margin. Notably, \model{}'s performance is superior to the generator-only model, illustrating the importance of the discriminators for generating more coherent summaries. Interestingly, on the more domain-specific AAN subset, our model is over 12\% better on ROUGE-L compared to the PGen baseline and 5.86\% better than the best extractive model. Our model also outperforms the strongest baselines on BERTScore. 


\begin{table*}[t]
\centering
\scalebox{0.8}{%
\begin{tabular}{l|rrr|rrr|rrr} \toprule
\multicolumn{1}{c}{\multirow{2}{*}{\textbf{Model}}} & \multicolumn{3}{c}{\textbf{AAN}} & \multicolumn{3}{c}{\textbf{CS}} & \multicolumn{3}{c}{\textbf{Bio}}\\
\multicolumn{1}{c}{}& \textbf{R-1} & \textbf{R-2} & \multicolumn{1}{c}{\textbf{R-L}} & \textbf{R-1} & \textbf{R-2} & \multicolumn{1}{c}{\textbf{R-L}} & \textbf{R-1} & \textbf{R-2} & \textbf{R-L} \\
\midrule
Coverage (Cov) & 41.29 & 12.09 & 37.14 & \textbf{38.57} & 10.81 & \textbf{35.11} & \textbf{35.86} & 8.41 & \textbf{32.56} \\
Order       & 41.20 & 12.20 & 37.11 & 38.50 & 10.87 & 35.10 & 35.66 & \textbf{8.46} & 32.39 \\
Adjacency (Adj)         & 40.97 & 12.46 & 36.70 & 37.44 & 10.67 & 33.86 & 34.89 & 8.45  & 31.57 \\
Factuality  & \textbf{41.67} & \textbf{12.65} & \textbf{37.23} & 38.23 & \textbf{11.03} & 34.64 & 35.46 & 8.41 & 31.96 \\
\bottomrule
\end{tabular}}
\caption{Comparison of different Co-opNet discriminators}
\label{table:auto_coop}
\end{table*}


\begin{table*}[!ht]
\centering
\scalebox{0.8}{%
\begin{tabular}{l|r|r|l|r|r} \toprule
    \multicolumn{3}{c}{\textbf{PGen vs. \model-Adj}} & \multicolumn{3}{c}{\textbf{Generator vs. \model-Adj}} \\ 
    \multicolumn{1}{c}{\textbf{Criteria}} & \multicolumn{1}{c}{\textbf{PGen}} & \multicolumn{1}{c}{\textbf{\model}} & \multicolumn{1}{c}{\textbf{Criteria}} & \multicolumn{1}{c}{\textbf{Generator}} & \textbf{\model}\\ \midrule
    Abstractiveness & 41.89           & \textbf{47.30}    & Abstractiveness &  20.41    & \textbf{38.10} \\
    Coherence       & 42.57           & \textbf{50.00}    & Coherence       &  23.81    & \textbf{34.01}\\
    Factuality      & 39.86           & \textbf{45.95}    & Factuality      &  22.98    &  \textbf{30.41} \\
    Overall         & 34.90           & \textbf{53.02}    & Overall         &  25.00    &  \textbf{31.08}  \\ \bottomrule
    
\multicolumn{6}{c}{} \\ \toprule
     \multicolumn{3}{c}{\textbf{PGen vs. \model-Fact}} & \multicolumn{3}{c}{\textbf{Generator vs. \model-Fact}} \\
    \multicolumn{1}{c}{\textbf{Criteria}} & \multicolumn{1}{c}{\textbf{PGen}} & \multicolumn{1}{c}{\textbf{\model}} & \multicolumn{1}{c}{\textbf{Criteria}} & \multicolumn{1}{c}{\textbf{Generator}} & \textbf{\model}\\ \midrule
    Abstractiveness & \textbf{51.02}  & 39.46             & Abstractiveness &  27.33    & \textbf{35.33} \\
    Coherence       & 43.92           & \textbf{50.00}    & Coherence       &  30.87    & \textbf{32.21}\\
    Factuality      & 43.84           & \textbf{48.63}    & Factuality      &  27.52    &  \textbf{32.21} \\
    Overall         & 43.54           & \textbf{50.34}    & Overall         &  30.87     & \textbf{32.21}  \\ \bottomrule
\end{tabular}}
\caption{Human Evaluation of Co-opNet Architectures (\% of judgements for each model)}
\label{table:human_gen}
\end{table*}


When we break down results for various \model~architectures (see Table \ref{table:auto_coop}), we find that the factuality and discourse role discriminators lead to the best performance in terms of ROUGE scores with the adjacency discriminator achieving lower performance on ROUGE than the base generator. However, as shown by Table \ref{table:auto_bert}, the adjacency discriminator outperforms the base generator when we consider BERTScore, a more contextual evaluation metric, indicating that this generator-discriminator combination selects summaries that capture the same linguistic patterns and meaning as reference summaries without directly copying.

%% file: experiments/human_eval.tex
\subsection{Human Evaluation} 
\label{ssec:human_eval}
\begin{table*}[t] 
\centering
    \small 
  \begin{tabular}{p{2.15cm}|p{12.55cm}}
    \multirow{4}{*}{\textbf{Gold}} & We investigate mutual benefits between syntax and semantic roles using neural network models, by studying a parsing->SRL pipeline, a SRL->parsing pipeline, and a simple joint model by embedding sharing. The integration of syntactic and semantic features gives promising results in a Chinese Semantic Treebank... \\ \midrule \midrule
    \multirow{3}{*}{\textbf{PGen}} & In this paper, we propose a novel approach to learn \textit{\textbf{syntactic and semantic role labeling models}} \textit{to semantic role labeling (wsd)}. \textit{In the first} neural network models induce non-linear \textit{feature features} from word and \textit{\textbf{part-of-speech (pos) parsing}}. We show that semantic features can be used to learn...  \\ \midrule \midrule
    \multirow{4}{*}{\textbf{Generator}} & \textit{\textbf{Syntax-semantic relations}} play a crucial role in natural language processing. \textit{In contrast,} \textit{\textbf{semantic role labeling (srl)}} models typically rely on parser output features to improve accuracy. In this work, we propose a joint \textit{\textbf{srl and syntactic parsing srl pipeline}} using the \textit{\textbf{chinese treebank}} \textit{(qiu et al., 2016)}...\\ \midrule \midrule
    \multirow{3}{*}{\textbf{\model{} (Adj)}}& In this paper, we explore the use of neural network models to jointly train \textit{\textbf{semantic role labelers}} and \textit{\textbf{parsers}} for \textbf{\textit{semantic role labeling (srl)}}. We first propose a simple neural \textit{\textbf{srl}} model that uses a neural \textbf{\textit{long shortterm memory (lstm)-based parser}} to represent the output of an \textit{\textbf{srl}} system...
  \end{tabular}
  \caption{Example of gold and generated abstracts from baseline Pointer Networks + Coverage~\cite{see2017get} (PGen) and two of our proposed models, Generator and \model, on the NLP scientific domain. Coherence issues and factual errors in generated abstracts are highlighted in \textit{italics}. We highlight correct terminology and transitional phrases that contribute to coherent flow by properly delineating sections of abstracts in \textbf{\textit{bold}} and \textbf{\textit{italics}}.}
  \label{table:big_wall_of_examples}
\end{table*}

Since coherence of generated text is difficult to measure with automatic metrics \cite{kilickaya2017re,Sun2019HowTC,clark3019sentence}, we conduct human evaluations to assess how the discriminator affects generation quality using pairwise model comparisons. 

\paragraph{Setup} We use four key criteria in all evaluations – abstractiveness, coherence, factuality and best overall quality, which we define as follows:

\begin{itemize}
    \item Abstractiveness $\rightarrow$ Which abstract rewords information from the introduction instead of directly copying from the introduction?
    \item Coherence $\rightarrow$ Which abstract is more structured, and presents a complete and coherent story about the work done in the paper?
    \item Factuality $\rightarrow$ Which abstract is more factually consistent, presenting the same information that appears in the introduction and not producing hallucinated information?
    \item Overall $\rightarrow$ Which abstract is better overall?
\end{itemize}


\noindent We conduct human evaluations on Amazon Mechanical Turk (AMT) considering 4 different abstractive baseline model variants over 100 randomly sampled AAN test set examples. Given a gold introduction, AMT evaluators are asked to compare a corresponding abstract generated from \model~against an abstract generated by a baseline or our generator model. To reduce bias, the ordering of generated abstracts are randomized and evaluators are not told that abstracts are machine-generated. 

Each abstract pair is judged by three unique annotators. For each criteria, we filter to 50 abstracts based on the amount of time AMT workers spent ($\geq$ 20 seconds) and inter-annotator agreement (at least $\frac{2}{3}$ of annotators should agree on which abstract is best). 
We also prime annotators to consider subtler aspects of discourse coherence by providing examples that capture good or bad narrative flow without complete text degeneration. 

We test the Co-opNet framework using both the factuality and adjacency discriminators, as these are the highest and lowest performing discriminator architectures in terms of automatic metrics on the AAN domain. We allow for ties, as \model{} and the generator baseline sometimes assign the highest probability to the same abstract, or generated abstracts in the candidate pool are high quality enough that there is little room for improvement. 

\paragraph{Results}

We find that \model~is preferred across all criteria for all comparisons, when we use the adjacency discriminator (see Table \ref{table:human_gen}). When using the the factuality discriminator, \model{} is superior to baselines in all cases except when compared on abstractiveness to the PGen model. 

In particular, human evaluators prefer \model~with the adjacency discriminator over baselines by over 8\% on the coherence metric and 18.12\% compared to PGen on overall quality. Notably, the adjacency discriminator encourages more abstractiveness in generated abstracts while still maintaining higher levels of factual consistency. We also find that \model~with the factuality discriminator improves coherence and overall quality in addition to factuality. However, \model~generations with the factuality discriminator were found to be more extractive than abstracts generated by PGen. 

As shown in Table \ref{table:big_wall_of_examples}, generations selected by the adjacency discriminator more closely match the distribution of abstracts, while the generator sometimes favors copying from the introduction at the loss of narrative structure. For example, the generator will select a summary that opens with ``\textit{we present a method for jointly solving penn treebank style empty category (e.g. figure 1)...}", while the adjacency discriminator selects a summary that opens with ``\textit{we present a method to jointly solve the problem of empty categories...}" and does not refer to a particular figure. Both summaries are faithful to the introduction, but the discriminator-selected summary makes more sense in the context of a paper abstract.

\input{experiments/examples.tex}

%% file: 7-related.tex
\section{Related Work}




\paragraph{Narrative Flow and Factuality}

Modeling coherent narrative flow remains a major challenge in the field of text generation, due to the need for accurate understanding of narrative structure \cite{christensen-etal-2013-towards,Nikolov2018DatadrivenSO,holtzman2018learning,Qin2019CounterfactualSR,KoncelKedziorski2019TextGF,Gabriel2021ParagraphLevelCT}. Early approaches to incorporating structure include integration of explicit discourse markers into automatic summarization \cite{alonso-i-alemany-fuentes-fort-2003-cohesion}. Recently proposed solutions include global-tracking of entities \cite{Kiddon2016GloballyCT,Bosselut2018DiscourseAwareNR,Mei2016CoherentDW}, as well as discourse-aware attention \cite{cohan2018discourse}. While there has been prior work on factual consistency \cite{Cao2018FaithfulTT,gao2019write,kryciski2019evaluating,zhang2019optimizing}, these works did not focus on scientific paper summarization. 

\paragraph{Neural Abstractive Summarization}
In the past, abstractive summarization models \cite{rush2015neural,Gehrmann2018BottomUpAS} have relied upon seq2seq encoder-decoder architectures \cite{sutskever2014sequence,xsum,elikyilmaz2018DeepCA}. 
Transformer models have emerged as a promising architecture for text generation and summarization \cite{Liu2018GeneratingWB,Hoang2019EfficientAO,Khandelwal2019SampleET,Zhang2019HIBERTDL}. While our model builds upon this work, it is, to our knowledge, the first transformer summarization framework to explicitly model narrative flow and scientific fact-checking across domains.

%% file: 8-conclusion.tex
\section{Conclusion}

In this work, we introduced \textit{Cooperative Generator-Discriminator Networks}, a framework for more coherent natural language generation with transformer language models through the integration of discriminators that encourage proper narrative flow and factual consistency. 
Through our analyses over scientific papers from \dataset~and AAN, we empirically showed that our framework selects generations that are more relevant and narratively coherent than previous approaches. 



%% file: 9-acknowledgments.tex
\section*{Acknowledgments}

We thank the anonymous reviewers, as well as Lianhui (Karen) Qin, Jungo Kasai, Rik Koncel-Kedziorski, Elizabeth Clark, Dave Wadden and Rowan Zellers for helpful feedback. We also thank the annotators who contributed to the human evaluations in this work. This research was supported in part by NSF (IIS-1524371), DARPA CwC through ARO (W911NF-15-1-0543), and Samsung AI Research. 

%% file: appendix.tex
\appendix
\section{Appendices}
\label{sec:appendix}

\subsection{Additional Implementation Details}

\subsection{Baselines}
\label{baselines}

For the sequence-to-sequence RNN model, a bi-LSTM is used to encode a given source article $a$ and a separate decoder LSTM produces the generated summary $g$. At each decoding time step, the decoder attends to all the context vectors produced by the encoder as well as the maintained state from the previous decoder tokens to produce the next token in the summary.

The Pointer-Generator (PGEN + Cov) model extends the base LSTM model (LSTM + Cov) to allow tokens to be copied from the input during generation. Baselines are trained for up to 40000 steps with a batch size of 16. Following previous work, we decode from these baselines using beam search with a beam size of 4. 

\subsection{Generator Model}
\label{disc-a}

We use the 345M parameter GPT-2 model. The model is trained to minimize the negative log likelihood of the next word $w_i$ given all preceding words: 
\begin{equation}
\mathcal{L}_{gen} = - \sum_{i=1}^{\vert a \vert + \vert s \vert} \log P(w_i|w_0,...w_{i-1})
\label{eq:train_obj}
\end{equation}

\noindent where $w_i$ is the $i^{th}$ token of our full input vector $X$, $a$ is our article and $s$ is our summary. At test time, $X$ only consists of the gold article and delimiter token $(a_1, ..., a_{\vert a \vert}, [\mathrm{SEP}]$) and we decode generated summaries $g$ starting from this input. 


During generation, we filter candidate summaries from the hypothesis generation pool that contain sentences longer than a fixed max length of 200 tokens, a clear sign of coherence deterioration. We use a candidate pool size of 30 for ATLAS and 20 for AAN. 

\subsection{Discriminator Training}
\label{disc-b}

\paragraph{Factuality Discriminator Details}

For the token-level classification model, we use the BERT base model with binary labels for whether or not a token should be included in a salient span. We predict for all spans in an abstract at once.

\paragraph{Order Discriminator Details} 

We set the max length of summaries considered by the order discriminator to be 10 sentences, truncating longer summaries. Given the max length of a summary, we have a fixed number of orderings |O| that can be scored. We calculate the final score from the order discriminator based on the unnormalized sum of scores from these orderings, S, and the following function $f_{n}$:
\begin{equation}
    f_{n}(S,|O|) = \frac{S-(-|O|)}{|O| - (-|O|)}
\end{equation}

\paragraph{Sentence Selection for Discriminator Models}

To train an adjacency discriminator model, we use a subset of adversarial and positive sentence pair examples extracted from the training set. The sentence pairs are extracted from gold abstracts containing at least five sentences using the following approach: For a randomly selected sentence $s_u$ from the abstract, we randomly select an adjacent sentence, $s_{u-1}$ or $s_{u+1}$, as a positive example and any nonadjacent sentence $s_{v \notin [u-1,u, u+1]}$ as a negative example.

\paragraph{Discriminator Performance}

We measure the performance of discriminator models using recall, precision, accuracy and F1. Table \ref{table:auto_disc} provides summary statistics of discriminator performance on the various discourse and factuality objectives. Discourse-Adj denotes the adjacency discriminators, while Discourse-Abs denotes the discourse role label prediction model \cite{Cohan2019PretrainedLM} and Factuality denotes the token saliency prediction model.
\begin{table}[!htb]
\centering
\scalebox{0.7}{%
\begin{tabular}{l|l|r|r|r|r} \toprule
     \multicolumn{1}{l}{\textbf{Model}} & \multicolumn{1}{c}{\textbf{Training Data}} & \multicolumn{1}{c}{\textbf{Prec}} & \multicolumn{1}{c}{\textbf{Rec}} & \multicolumn{1}{c}{\textbf{F1}} & \textbf{Acc} \\ \midrule
     Discourse-Adj & \dataset-AAN  & 86.05 & 85.25  & 85.65 & 86.81 \\
     Discourse-Adj & \dataset-Bio  & 90.30 & 93.44  & 91.84 & 92.32\\
     Discourse-Abs & CSAbstruct & 88.99 & 89.09  & 89.04 & 89.00\\ 
     Factuality    &  SciFact   & 73.70 & 70.50  & 72.10 & 75.70 \\  \bottomrule
\end{tabular}}
\caption{Automatic Evaluation of discriminator architectures}
\label{table:auto_disc}
\end{table}

\subsection{Details on Model Performance}

Automatic results for Co-opNet selection were given using a context size of 800 tokens for the input, while a context size of 800 characters was used to select Co-opNet summaries for the human eval. The automatic results for the summaries used in the human eval were lower than the ones using the longer context size. Using a smaller context size leads to faster and more efficient Co-opNet selection (less memory usage), but slightly lower overall automatic performance (while maintaining the same ordering in terms of highest and lowest scores for Co-opNet variants on ROUGE). 

\subsection{Comparison of Datasets}
\label{datasets}

\begin{table*}[t]
\begin{center}
\resizebox{.8\textwidth}{!}{%
\begin{tabular}{  l c r r r} 
\toprule
Dataset  & Narrative Flow? & \# Summaries & Avg \# Sents & Avg \# Words \\
\midrule
 XSum \cite{xsum}& \xmark & 226,711 & 1.00 &   23.26      \\ 
Newsroom \cite{newsroom} & \xmark &  1,321,995 & 1.45  &   26.70   \\ 
CNN \cite{cnndm} & \xmark & 92,579 &  3.59 & 45.70\\
DailyMail \cite{cnndm} & \xmark & 219,506 &  3.86 &  54.65\\
\dataset & \cmark & 472,493 & 6.11 & 150.85\\
AAN & \cmark & 11,890 & 5.03 & 106.76\\
\bottomrule
\end{tabular}
}
\caption{Statistics of gold summaries in different summarization datasets.}
\label{table:sentstats}
\end{center}
\end{table*}

We removed duplicates and articles without abstracts from AAN. From this subset, we extract introduction and abstract pairs. 

\subsection{Additional Analysis} 

\paragraph{Comparison with Gold Summaries} 

To obtain an upper-bound comparison for the human evaluation and verify the effectiveness of our human evaluation pipeline for judging the quality of abstracts, we used the same intro-abstract pairs and Mturk annotation framework as the model comparison to conduct a Turing-style evaluation. In this evaluation, we presented a Co-opNet (adj) generated abstract and a gold abstract to the annotators in a random ordering without noting whether either of the abstracts were human-written or machine-generated. We found that annotators consistently selected the gold abstract over the machine-generated abstract when considering factuality and coherence, though they found the machine-generated abstracts to be slightly more abstractive. We provide the results for this full evaluation in Table \ref{table:turing}. 

\begin{table*}[!ht]
\centering
\scalebox{1}{%
\begin{tabular}{l|r|r}
    
\multicolumn{3}{c}{} \\ \toprule
 \multicolumn{3}{c}{\textbf{Gold vs. \model-Adj}} \\
    \multicolumn{1}{c}{\textbf{Criteria}} & \multicolumn{1}{c}{\textbf{Gold}} & \multicolumn{1}{c}{\textbf{\model}} \\ \midrule
    Abstractiveness & 47.37  & \textbf{52.63} \\
    Coherence       & \textbf{66.67}           & 31.06 \\
    Factuality      & \textbf{66.92}         & 32.33  \\
    Overall         & \textbf{61.36}          & 32.33     \\ \bottomrule
\end{tabular}}
\caption{Human Evaluation of Co-opNet Architectures (\% of judgements for each model)}
\label{table:turing}
\end{table*}